%% file: main.tex
\documentclass[11pt]{article}
\usepackage{coling2020}
\usepackage{times}
\usepackage{url}
\usepackage{latexsym}

\usepackage{soul} 

\newcommand\myscale{.6}
\usepackage{amsfonts}
\usepackage{todonotes}

\newenvironment{customlegend}[1][]{%
        \begingroup
        \csname pgfplots@init@cleared@structures\endcsname
        \pgfplotsset{#1}%
    }{%
        \csname pgfplots@createlegend\endcsname
        \endgroup
    }%
    \def\addlegendimage{\csname pgfplots@addlegendimage\endcsname}
\usepackage{tikz}
\usepackage{pgfplots}
\usepgfplotslibrary{fillbetween}

\colingfinalcopy 

\title{Context-based Transformer Models for Answer Sentence Selection}

\author{Ivano Lauriola \\
  $^1$ University of Padova, Dept of Mathematics \\
  $^2$ Fondazione Bruno Kessler \\
  {\tt ivano.lauriola@phd.unipd.it} \\\And
  Alessandro Moschitti \\
  Amazon Alexa \\
  {\tt email@domain} \\}

\date{}

\begin{document}
\maketitle

\begin{abstract}
An important task for the design of  Question Answering systems is the selection of the sentence containing (or constituting) the answer from documents relevant to the asked question.
Most previous work has only used the target sentence to compute its score with the question as the models were not powerful enough to also effectively encode additional contextual information.
In this paper, we analyze the role of the contextual information in the sentence selection task, proposing a Transformer based architecture that leverages two types of contexts, local and global.
The former describes the paragraph containing the sentence, aiming at solving implicit references, whereas the latter describes the entire document containing the candidate sentence, providing content based information.
The results on three different benchmarks show that the combination of local and global contexts in a Transformer models significantly improves the accuracy in Answer Sentence Selection.
\end{abstract}

\section{Introduction}

Answer Sentence Selection (AS2) is the task of identifying sentences that contain the answer to a given question in documents relevant to the question, e.g., retrieved by a search engine. Neural models have highly improved the accuracy on such task and recent approaches, e.g., using Transformer architectures as shown in \cite{garg2019tanda}, have shown an impressive accuracy.
%
%
Although AS2 models seem hard to improve, we note that most previous work does not exploit contextual information in addition to the candidate sentence. For example, \cite{tan2017context} used context in a hierarchical gated recurrent networks but their accuracy is 10-12 points below the state of the art \cite{garg2019tanda} (as measured on the same exact dataset). Thus we cannot infer that context is really useful in state-of-the-art models for AS2.
However, we do know that some sentences contain ambiguities that cannot be solved without using references to the context outside of the target sentence.

\begin{table}[htb]
\centering
\small
\begin{tabular}{|l|p{.7\linewidth}|}
	\hline
	Question	& \textbf{When was Lady Gaga born?}	\\
	\hline
	Prev. & Lady Gaga is an American singer, songwriter, and actress. \\
	Target & \textcolor{red}{She was born in 1986.} \\
	Next & Both of her parents have Italian ancestry, and\dots\\
	\hline
\end{tabular}
\caption{An example of correct answer sentence requiring larger local context to be selected.}
\label{tab:example}
\end{table}

\begin{table}[htb]
\centering
\small
\begin{tabular}{|l|p{.7\linewidth}|}
	\hline
	Question	& \textbf{Which role did Bradley Cooper play with Lady Gaga?}	\\
	\hline
	doc. title & Avengers: endgame - Movie plot \\
	sentence & Rocket Raccoon was voiced by Bradley Cooper.\\
	\hline
	doc. title & A star is born - Movie plot \\
	sentence & \textcolor{red}{Jackson "Jack" Maine (Bradley Cooper), a famous country rock singer...} \\
	\hline
	doc. title & American sniper - Movie plot \\
	sentence & Chris Kyle, the leading actor, was played by Bradley Cooper.\\
	\hline
\end{tabular}
\caption{Each of the three sentences can be a correct answer. Only the global document information, e.g., the title and the link between document concepts, allows us to select the correct sentence.}
\label{tab:globalexample}
\end{table}

For example, Table~\ref{tab:example} shows a simple question asking for the birthdate of \emph{Lady Gaga}. The answer is the middle sentence contained in a paragraph of three sentences. Clearly, an AS2 classifier cannot select the middle sentence with high reliability since the sentence does not reveal that \emph{she} refers to \emph{Lady Gaga}.
On the other hand, AS2 is effective as it targets just one sentence at a time: selecting an entire paragraph to be sent to the users, often provides them with too much irrelevant information\footnote{Of course, a solution based on a summarization approach would be optimal but poses complicated challenges, which have prevented to obtain better solutions than AS2 (to our knowledge).}.
A further example is described in Table~\ref{tab:globalexample}, where the question asks for the role of \emph{Bradley Cooper} in a specific movie. 
Each of the three sentences may reasonably be a correct answer. Also, the title of the movie is not enough to select the right answer and can be too far from the local context window.
However, \emph{``A star is born - Movie plot''} is the only document that contains references to \emph{Lady Gaga}. This related information allows us to recognize the correct answer.
Thus, to improve AS2, we would need local context, e.g., to solve the pronoun \emph{she}, and global information from the whole document to connect multiple concepts, e.g., the movie title with Lady Gaga.


In this paper, we propose to model local and global contexts for AS2 by ensembling multiple sentences in Transformer networks \cite{vaswani2017attention} and Bag-of-Word (BOW) features, respectively. 
More specifically, we consider candidates as a triplet ($s_{i-1}$, $s_i$, $s_{i+1}$), where $s_i$ is the target answer sentence and $s_{i-1}$ and $s_{i+1}$ are the preceding and the next sentence of $s_i$, respectively.
We integrate this triplet in Transformer architectures by using one single RoBERTa \cite{Liu2019RoBERTaAR} model encoding the three sentences in three embeddings. Then, we add document-level BOW representation is the classification layer.
We tested our models on three different datasets, Google NQ and SQuAD adapted for the AS2 task, as well as the well-known WikiQA, comparing with the very recent state of the art in AS2 \cite{garg2019tanda}. The results clearly show that local and global contexts can improve AS2 models.

\section{Related Work}
\label{sec:architecture}




Question Answering (QA) research manly regards two tasks: (i) AS2, which, given a question and a set of answer sentence candidates, consists in selecting sentences (e.g., retrieved by a search engine) that correctly answer the question; and (ii) Machine Reading (MR) or reading comprehension \cite{DBLP:journals/corr/ChenFWB17}, which, given a question and a reference text, involves finding an exact text span answering it.  AS2 research originated from the TREC competitions \cite{wang-etal-2007-jeopardy}, which target large databases of unstructured text.

Neural models have significantly contributed to both areas with new techniques, e.g.,  \cite{DBLP:journals/corr/WangJ16b,DBLP:journals/corr/abs-1904-07531,DBLP:journals/corr/abs-1901-04085}. 
In particular, recent approaches to neural language models, e.g., ELMO \cite{DBLP:journals/corr/abs-1802-05365}, 
BERT \cite{DBLP:journals/corr/abs-1810-04805}, RoBERTa  \cite{DBLP:journals/corr/abs-1907-11692},  XLNet \cite{DBLP:journals/corr/abs-1901-02860} have led to major advancements in several NLP subfields.
These methods capture dependencies between words and their compounds by pre-training neural networks on large amounts of data. Interestingly, the resulting models can be easily applied to different tasks by fine-tuning them on the target training data.
The impact of such methods on AS2, also thanks to transfer learning, is impressive. For example, \cite{garg2019tanda} exceeded the state of the art by 50\% (relative error reduction) on WikiQA \cite{yang2015wikiqa} and TREC-QA \cite{wang-etal-2007-jeopardy} datasets.
However, the proposed Transformer methods only focus on the similarity between the question and the candidate sentence pairs, without taking any additional information into account.

Contextual information was already introduced in neural networks for solving AS2, e.g., \cite{tan2017context}, by combining question/answer pairs with context information, selected by applying a similarity between question and document sentences.
Our model is built with state-of-the-art Transformer models for AS2, which we improve. We also improve the results from \cite{tan2017context} by a huge margin (+12\% on  WikiQA and +5\% on SQuAD). Finally, our approach is more modular and can be easily extended with additional context definitions.

\subsection{Transformer model for AS2}
\label{sec:related-transformer}
The Transformer 
is a popular neural network designed to learn language models, e.g., dependencies between words, in a context. Transformer models have recently shown to have remarkable impact on AS2 when used as ranker \cite{shao2019transformer,garg2019tanda,kumar2019improving}. 
Besides architectural definitions, Transformer models take advantage for extensive pre-training the language model (i.e., the q/a pair representation) on large-scale corpora by using the masked language model and next sentence prediction \cite{DBLP:journals/corr/abs-1810-04805}.
In those works, the question answer candidate pairs are codified as a joint sequence of tokens with specialized delimiters and separators, i.e.,
\emph{[CLS] $q^1\dots q^n$ [SEP] $s^1\dots s^m$ [EOS]}, where $x^j$ defines the $j$-th token of the sequence $x$.
\emph{[CLS]}, \emph{[SEP]}, and \emph{[EOS]} are special tokens used to mark the beginning of the sequence, the separation between question and candidate answer tokens, and the end of the text.
Several Transformer blocks are applied and then the representation associated with \emph{[CLS]} is used in a linear fully-connected layer to to compute the  final score associated with the question/answer pair.
The same concepts can be applied to RoBERTa or other pre-trained Transformer models.


\section{Contextual Transformer for AS2}
\label{sec:contextual}

To our knowledge no Transformer model for AS2 models use context, except for the information on the sentences. This is critical as a sentence may contain references to other part of the text and to external entities (see the example in Table~\ref{tab:example}).
We enhance the standard Transformer model for AS2 considering two types of contexts: local and a global.
The former aims at solving coreferences between the answer constituents by relating the candidate sentence to its neighborhood (typically corresponding to the paragraph containing the sentence). The latter introduces information concerning the topics and concepts of the document containing the target sentence.

\subsection{Local context}
\label{sec:local_ctx}
\begin{figure}[htb]
	\centering
	\includegraphics[width=.6\linewidth]{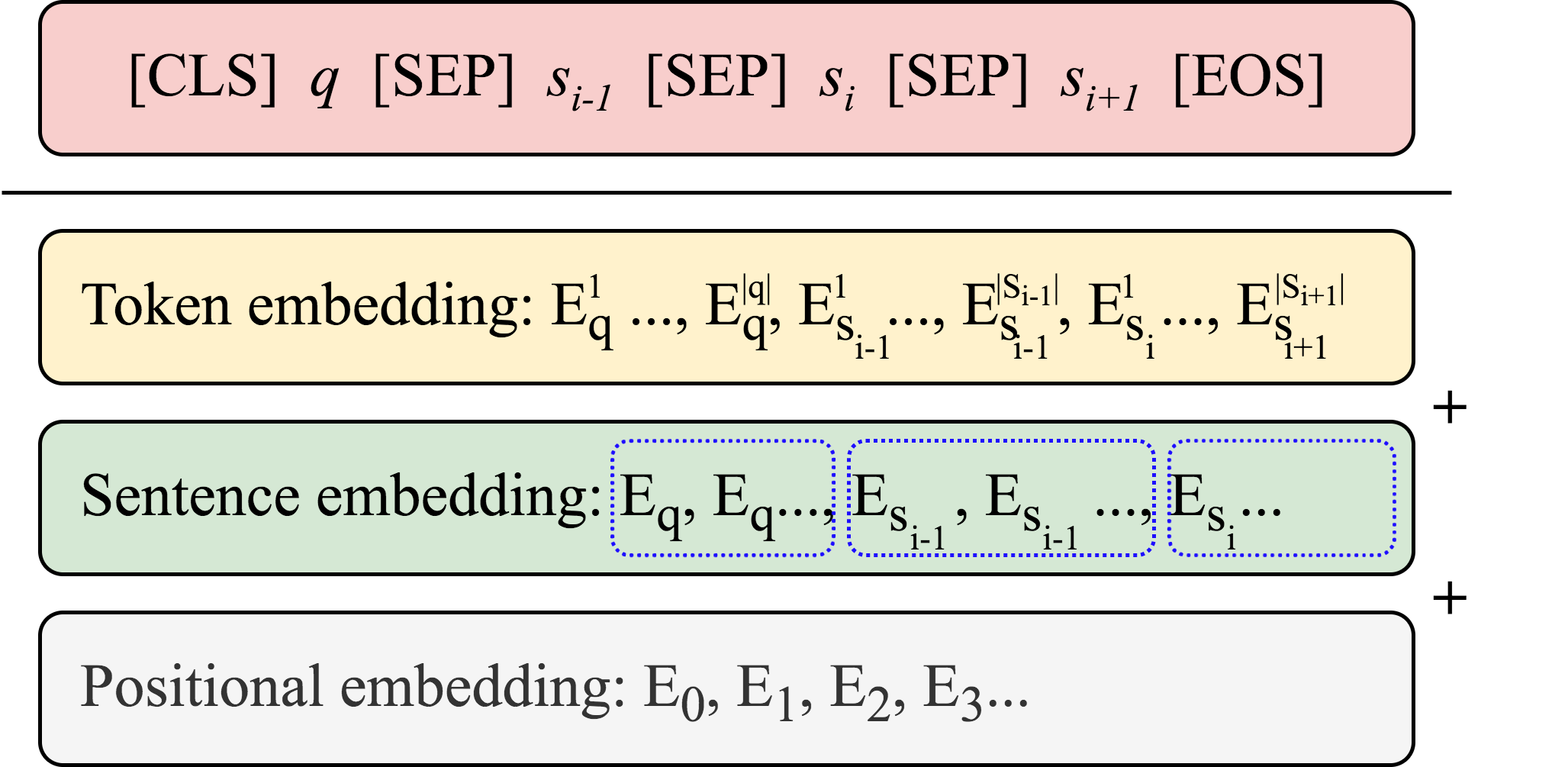}
	\caption{BERT/RoBERTa input sequences.}
	\label{fig:input}
\end{figure}

Given the target answer sentence candidate, $s_i$, we extend the AS2 model using the preceding, $s_{i-1}$, and the following, $s_{i+i}$, sentences. The (local) contextual ranker $r_\textsc{l}$ takes four elements as input and provides the following answer:
$
a_\textsc{l} = \arg\max_{s_i \in \mathcal{S}(q)} {r_\textsc{l}}(q, s_{i-1},s_i,s_{i+1}),
$
where $\mathcal{S}(q)$ is the set of relevant sentences for the question $q$ and $r_\textsc{l}$ is our ranking function, 
To implement ${r_\textsc{l}}$ in the RoBERTa model, the input sequence becomes \emph{[CLS] $q$ [SEP] $s_{i-1}$ [SEP] $s_i$ [SEP] $s_{i+1}$ [EOS]}.
Additionally, RoBERTa encodes each input word by using three pieces of information: the token, the sentence, and the positional embeddings.
The first is a standard word-embedding. The positional embedding describes a token as a function of its position into the sequence. Finally, the sentence embedding defines a token as a function of the sentence that contains it.
The sentence embedding helps the model to distinguish between different input sentences: it can be seen as a particular word embedding of size four, one entry for each element of the input tuple, $(q, s_{i-1},s_i,s_{i+1})$.
This embedding plays a crucial role in our model to learn that 
the instance label is exclusively associated with the middle sentence. The three embeddings are then summed to produce the final representation of the sentences to be fed as input to the Transformer.
This process is visually described in \figurename~\ref{fig:input} (see dashed squares).

\subsection{Global context}
\label{sec:global_ctx}
Our global context describes the document content rather than the structure of the paragraph containing the answer. We define a global ranker, $r_\textsc{g}$ as
$
a_\textsc{g} = \arg\max_{s_i \in \mathcal{S}(q)} {r_\textsc{g}}(q,s_i,{d}(s_i))
$, where $d(s_i)$ is the document containing $s_i$.
A simple representation of ${d}(s_i)$ is the \emph{bag-of-words} (BOW) model, for which, we use the same dictionary used by RoBERTa model. We combine ${d}(s_i)$ with the \emph{[CLS]} representation (vector concatenation) before applying our ranking function.
To make the two vectors comparable (in terms of dimensions), we apply a random projection to the original BOW representation, thus mapping the document into a vectorial space with the same dimension\footnote{We also normalize the projection to prevent scaling issues.} of \emph{[CLS]} (i.e., 768).

\subsection{Combined context}
\begin{figure}[htb]
	\centering
	\includegraphics[width=.5\linewidth]{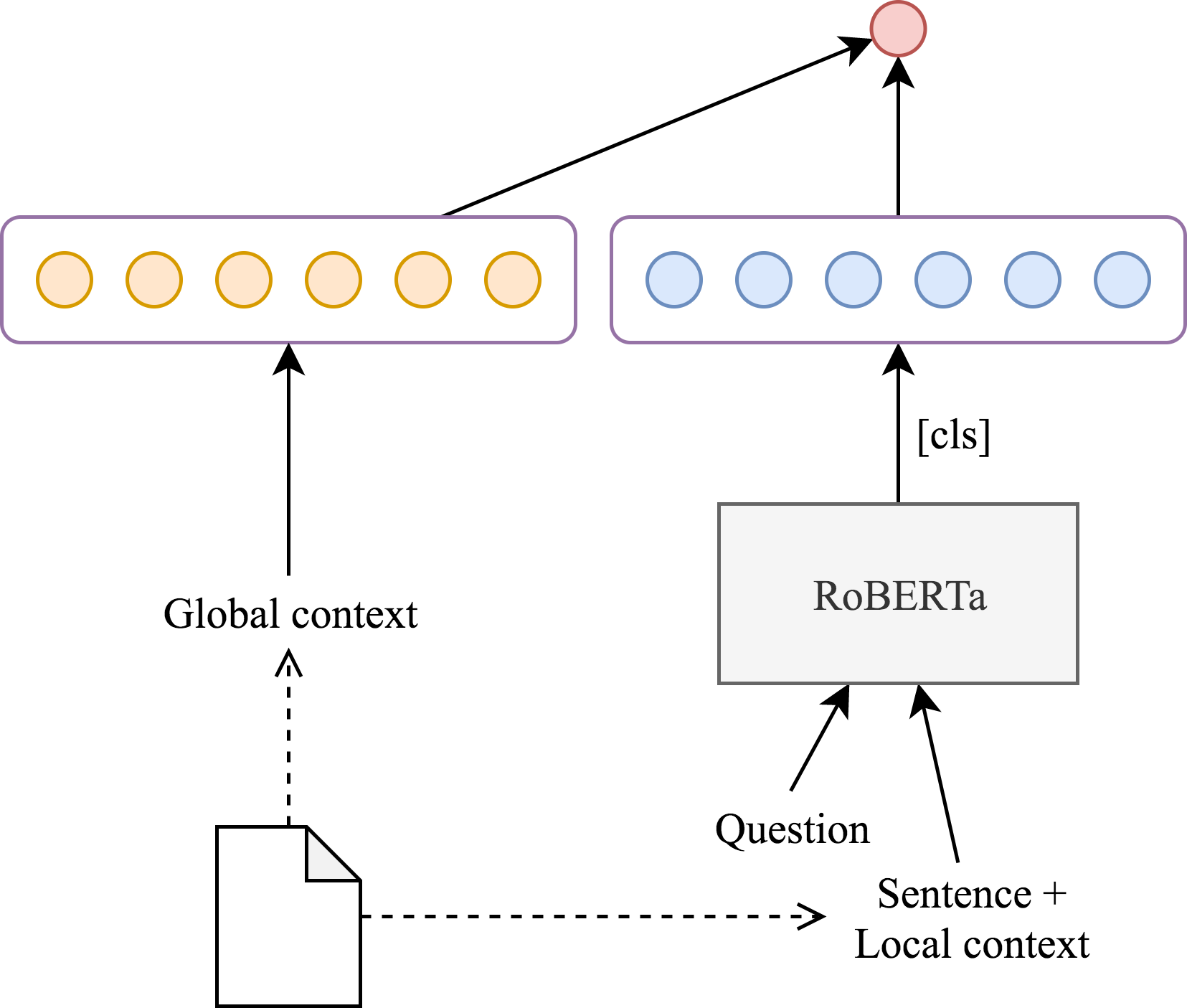}
	\caption{The proposed architecture that combines global and local contextual information.}
	\label{fig:architecture}
\end{figure}

Local and global contexts contain different information, thus we also combined them together in the model,
\textsc{Dual-CTX}. This is a RoBERTa model that receives the question and the candidate sentence with local context. The output of the Transformer is then combined with the global representation.
The architecture is modular and extensible, local and global feature extraction modules can be easily exchanged. 
However. the extensive evaluation of different context combinations and strategies is beyond the scope of this paper.
A general schema of the architecture is shown in \figurename~\ref{fig:architecture}.

\section{Empirical assessment}
We carried out comparative experiments to evaluate the local and global contexts and their combination.


\subsection{Corpora} 
We used three AS2 corpora, ASNQ, SQuAD, and WikiQA, to empirically assess the proposed contextual architecture.

\textbf{ASNQ}, Answer Sentence Natural Question \cite{garg2019tanda} is a large-scale open domain corpus for AS2. The corpus is built by transforming the recently proposed Natural Question (NQ) dataset \cite{NQ-2019} corpus from MR into AS2. In short, the corpus consists of 57,242 distinct natural questions for training, and 2,672 for development. For each question,  candidate answers have been extracted from a single Wikipedia page. 
The corpus contains 21,307,630 question/answer pairs, with an average of 356 answer candidates per question.
	
\textbf{WikiQA} \cite{yang2015wikiqa} is an open-domain corpus containing queries sampled from Bing logs. Based on the user clicks, the questions have been associated with a Wikipedia page (only the summaries were used).
We used the clean setting for which only questions having at least one good and one wrong answers are considered. The resulting corpus consists of 2,118 training, 126 development, and 243 test questions, with about 10 candidate answers per question on average.
We merged the dev. and test sets as they are too small to derive reliable results from each of them individually. Overall, we have 2117 questions and 20374 question/answer pairs.
	
\textbf{SQuAD 1.1}, Stanford Question Answering Dataset \cite{rajpurkar2018know},  is a large-scale corpus consisting of questions crowdsourced on a set of  20,000 Wikipedia articles. The dataset was designed for MR. We transformed it into a corpus for AS2 task, by applying the same procedure described by \cite{garg2019tanda}. In short, we split each input paragraph in sentences and labelled those containing the annotated answers as correct candidates, 
and all the others as negative candidates.
After this preprocessing, our corpus contains 87,355 questions and 448,108 question/answer pairs.
Please note that the results presented in this paper are not directly comparable to the SQuAD leaderboard\footnote{\url{https://rajpurkar.github.io/SQuAD-explorer/}}.


%

\begin{figure}[t]
	\centering
	\includegraphics[width=.7\linewidth]{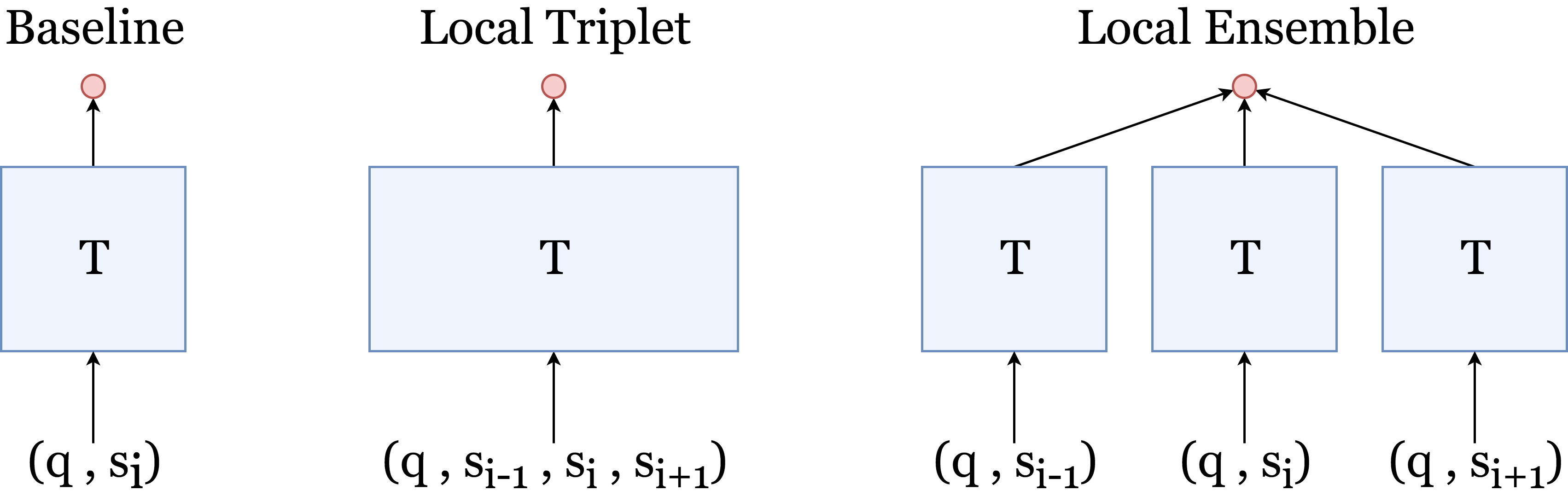}
	\caption{Our three approaches to encode local context: a simple Transformer with question/answer pairs (left), the contextual multi-sentence architecture (center), and the ensemble of Transformers (right).}
	\label{fig:architectures}
\end{figure}

\subsection{Models}

We implemented our methods with RoBERTa pre-trained models, using the shared checkpoint \cite{wolf2019transformers}. 
We fine-tuned the checkpoint on our data by using (i) the Adam optimizer set with the warmup linear scheduler and a learning rate peak of $1e-6$; (ii) the binary cross-entropy loss; (iii) a batch size of 64 examples on a single GPU to train on WikiQA and SQuAD; and (iii) a batch size of 512 examples distributed on 8 GPUs to train on the ASNQ corpus (which is much larger).
We used the official dev. set to derive the results, thus we set the parameters, i.e., learning rate, scheduler, and batch size, on a small portion of the training set (as our dev.~set). 
We train and test our models on SQuAD and WikiQA four times and take the average results to account for their variability.
Finally, we also used the models generated with TANDA (transfer and adapt) approach \cite{garg2019tanda} for WikiQA. The authors apply a first fine-tuning on ASNQ and then a second fine-tuning on the target data. TANDA is the current state of the art, 7-10 points better than any other approach on WikiQA.

We encode local context as depicted in \figurename~\ref{fig:architectures}:

\mbox{$\bullet$ } Transformer: the Transformer model for AS2 introduced in Sec.~\ref{sec:related-transformer}. It receives the question/answer pair as input without any context.

\mbox{$\bullet$ } Local Triplet (\textsc{Loc\_t}): the proposed Transformer-based method described in Sec.~\ref{sec:local_ctx}, which relies on three different sentences, i.e., the previous, the target, and the next;

\mbox{$\bullet$ } Local Ensemble (\textsc{Loc\_e}): an ensemble of three Transformer models encoding the three pairs, q/$s_{i-1}$, q/$s_{i}$, and q/$s_{i+1}$ and a final linear layer fed with the concatenation of the \emph{[CLS]} embeddings of the three models. The latter do not share their weights except those from \emph{[CLS]}. The ensemble is a more expensive approach.

The baseline models for encoding global context are:

\mbox{$\bullet$ } Global BOW (\textsc{Glob\_b}): the global context described in Sec.~\ref{sec:global_ctx} consisting of a simple Transformer model with a (compressed) BOW feature set on the top;

\mbox{$\bullet$ } Global Embedding (\textsc{Glob\_e}): a document embedding constituted by the average of the embeddings derived from all document sentences. We extract the sentence embedding using RoBERTa fine-tuned on ASNQ. We concatenate the average with the \emph{[CLS]} representation output by the AS2 Transformer model. 

%

\input{charts_local.tex}

\section{Results}
We tested different context models on three different datasets using the state of the art in AS2 as our baseline, i.e., the transformer model made available in \cite{garg2019tanda}. The latter improves 7-10 points all previous AS2 models on WikiQA and TREC-QA datasets.

\subsection{Local context}
\figurename~\ref{fig:res} shows the Mean Average Precision (MAP) and the Precision at 1 (P@1) for each epoch for the Transformer, \textsc{Loc\_t}, and \textsc{Loc\_e} models.
The plots show two main results: first, the superior accuracy of \textsc{Loc\_t} is evident on all corpora, demonstrating that the local context has a positive impact on the AS2 model accuracy. 
Additionally, the performance of \textsc{Loc\_e} method shows that the mere use of more information is not sufficient: its arrangement into the model is fundamental.
Indeed, the simple aggregation of the three summarized context vectors seems not able to capture sentence dependencies: disarranged information produces noise, with a consequent drop in performance.

Next, we used an MR Transformer \cite{wolf2019transformers} to implement a sentence selector model. Our MR approach achieves 0.881 of F1 score on MR task (showing competitive results on the SQuAD leaderboard with respect to single models).
Then, we simply select the sentence from which the MR extracts the answer span to solve the AS2 task on SQuAD: the model achieves a P@1 of 0.952. 
\figurename~\ref{fig:res} shows that such model (straight line) is comparable to our baseline (single Transformer models), whereas \textsc{Loc\_t} achieves better performance.
This is a loose comparison but it suggests that our approach may be applied to develop new MR methods.

\input{charts_global}

\subsection{Global context}

\figurename~\ref{fig:res_global} shows the MAP and the P@1 achieved by the simple Transformer and the two global models, i.e., \textsc{Glob\_b} and \textsc{Glob\_e}.  We also report the results of the combined model, which includes local and global contexts.
Finally, we evaluated the models when applied to WikiQA without the TANDA approach, showing their behavior in a scenario, when such approach cannot be used, i.e., there is no a large and general data for a first fine-tuning step.

The figure shows that both global methods, i.e., BOW and document embedding, improve the standard model both on WikiQA and SQuAD. We did not apply \textsc{Glob\_B} and \textsc{Glob\_E} to ASNQ as the training has a very large computational cost. This means that we cannot apply TANDA to WikiQA with such context. 
In any case, the global context produces accuracy increase on WikiQA and SQuAD (w/o TANDA).  
Concerning the combined model, \textsc{Dual-CTX} improves the overall performance on WikiQA (w/o TANDA) and SQuAD.
It does not improve the MAP of \textsc{Loc\_t} on WikiQA when TANDA is used, but P@1 receives an interesting boost.
This result provides evidence that global and local features describe different (and potentially orthogonal) information\footnote{We used BOW in the \textsc{Dual-CTX} rather than the document embedding for computational reasons.}. 

\section{Conclusion}
AS2 is an important sub-task of Question Answering, which provides an effective solution for the design of automated QA systems.
Usually, models for AS2 consider only the question and the candidate answer sentence, without taking the context into account.
In this paper, we define two types context, local and global. 
The former tries to solve implicit references in a candidate sentence, and it consists of the previous and successive sentence of a candidate answer.
Conversely, the global context injects document related information, such as the main content and topics.
We proposed Transformer based architectures that leverages the different context for AS2.
Our empirical assessment shows a remarkable improvement of the proposed approach on three different AS2 datasets, i.e., ASNQ, WikiQA, and SQuAD 1.1, adapted for AS2. We will release the contextualized checkpoints and the SQuAD adaption for AS2\footnote{The link will be available after the reviewing process}.

Interesting future extensions of our work regard the extraction of features from the entire rank of documents retrieved for a question. Clearly, learning to rank features can also improve the selection of answer sentences.

\bibliographystyle{coling}
\bibliography{coling2020}

\end{document}

%% file: charts_local.tex
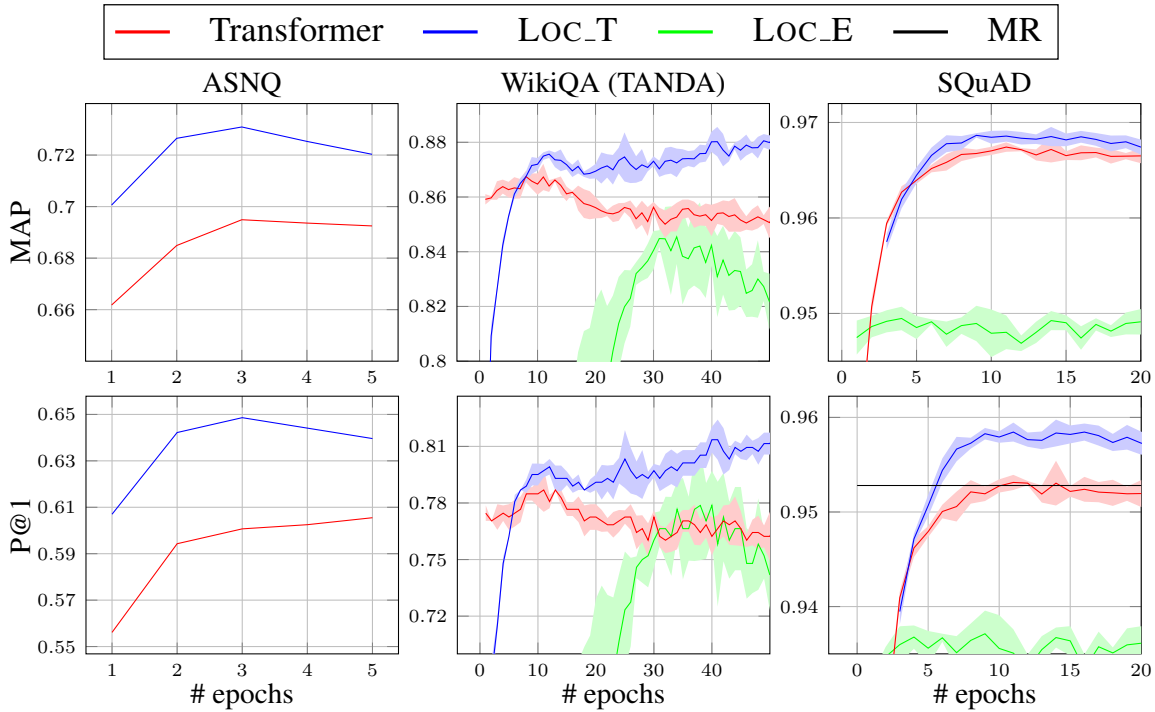
\begin{figure}[t]
\centering

\resizebox{.8\linewidth}{!}{
\begin{tikzpicture}
        \begin{customlegend}
        	[scale=\myscale,
        	legend columns=4,legend style={align=center,column sep=2ex},
        	legend entries={Transformer, \textsc{Loc\_T}, \textsc{Loc\_E}, MR}]
        \addlegendimage{red, line width=1pt}
        \addlegendimage{blue, line width=1pt}
        \addlegendimage{green, line width=1pt}
        \addlegendimage{black, line width=1pt}
        \end{customlegend}
\end{tikzpicture}
}

\begin{tikzpicture}
	\begin{axis}[
	        name=plotLeft,
			scale=\myscale,
			ylabel={MAP},
			ytick={0.66, 0.68, 0.70,0.72},
			grid=both,
			xtick={1,2,3,4,5},
			ymin=0.64,
			ylabel near ticks,
			tick label style={font=\scriptsize},
			ylabel style={yshift=-1ex},
			title={ASNQ},
			title style={yshift=-1.5ex,},
		]
		\addplot [red]
		    table[x=epo, y=map,] {results/asnq_base.txt};
		\addplot [blue]
		    table[x=epo, y=map] {results/asnq_tuples.txt};
		    
	\end{axis}
	\begin{axis}[
	        name=plotCenter,
			at=(plotLeft.right of south east), anchor=left of south west,
			scale=\myscale,
			ymin=0.8,
			xtick={0,10,20,30,40},
			xmax=50,
			grid=both,
			ylabel near ticks,
			tick label style={font=\scriptsize},
			xlabel style={yshift=1ex},
			title={WikiQA (TANDA)},
			title style={yshift=-1.5ex,},
		]
%


		\addplot [green]
		    table[x=epo, y=map,] {results/wikiqa_siamese_tanda.txt};
		\addplot [name path=upper_wst,draw=none] table[x=epo,y expr=\thisrow{map}+\thisrow{map_std}] {results/wikiqa_siamese_tanda.txt};
		\addplot [name path=lower_wst,draw=none] table[x=epo,y expr=\thisrow{map}-\thisrow{map_std}] {results/wikiqa_siamese_tanda.txt};
		\addplot [fill=green!20] fill between[of=upper_wst and lower_wst];
		
		\addplot [red]
		    table[x=epo, y=map,] {results/wikiqa_baseline_tanda.txt};
		\addplot [name path=upper_wbt,draw=none] table[x=epo,y expr=\thisrow{map}+\thisrow{map_std}] {results/wikiqa_baseline_tanda.txt};
		\addplot [name path=lower_wbt,draw=none] table[x=epo,y expr=\thisrow{map}-\thisrow{map_std}] {results/wikiqa_baseline_tanda.txt};
		\addplot [fill=red!20] fill between[of=upper_wbt and lower_wbt];
		
		\addplot [blue]
		    table[x=epo, y=map] {results/wikiqa_tuples_tanda.txt};
		\addplot [name path=upper_wtt,draw=none] table[x=epo,y expr=\thisrow{map}+\thisrow{map_std}] {results/wikiqa_tuples_tanda.txt};
		\addplot [name path=lower_wtt,draw=none] table[x=epo,y expr=\thisrow{map}-\thisrow{map_std}] {results/wikiqa_tuples_tanda.txt};
		\addplot [fill=blue!20] fill between[of=upper_wtt and lower_wtt];
	\end{axis}
	\begin{axis}[
	        name=plotRight,
			at=(plotCenter.right of south east), anchor=left of south west,
			scale=\myscale,
			ymin=.945,
			grid=both,
			xtick={0,5,10,15,20},
			xmax=20,
			ylabel near ticks,
			tick label style={font=\scriptsize},
			xlabel style={yshift=1ex},
			title={SQuAD},
			title style={yshift=-1.5ex,},
		]
		
		\addplot [green]
		    table[x=epo, y=map,] {results/squad_siamese.txt};
		\addplot [name path=upper_ss,draw=none] table[x=epo,y expr=\thisrow{map}+\thisrow{map_std}] {results/squad_siamese.txt};
		\addplot [name path=lower_ss,draw=none] table[x=epo,y expr=\thisrow{map}-\thisrow{map_std}] {results/squad_siamese.txt};
		\addplot [fill=green!20] fill between[of=upper_ss and lower_ss];
		
		\addplot [red]
		    table[x=epo, y=map,] {results/squad_baseline.txt};
		\addplot [name path=upper_sb,draw=none] table[x=epo,y expr=\thisrow{map}+\thisrow{map_std}] {results/squad_baseline.txt};
		\addplot [name path=lower_sb,draw=none] table[x=epo,y expr=\thisrow{map}-\thisrow{map_std}] {results/squad_baseline.txt};
		\addplot [fill=red!20] fill between[of=upper_sb and lower_sb];

		\addplot [blue]
		    table[x=epo, y=map] {results/squad_tuples.txt};
		\addplot [name path=upper_st,draw=none] table[x=epo,y expr=\thisrow{map}+\thisrow{map_std}] {results/squad_tuples.txt};
		\addplot [name path=lower_st,draw=none] table[x=epo,y expr=\thisrow{map}-\thisrow{map_std}] {results/squad_tuples.txt};
		\addplot [fill=blue!20] fill between[of=upper_st and lower_st];
		
%
	\end{axis}
\end{tikzpicture}


\begin{tikzpicture}
	\begin{axis}[
	        	name=plotLeft1,
	        	scale=\myscale,
			xlabel={\# epochs},
			ylabel={P@1},
			grid=both,
			xtick={1,2,3,4,5},
			ytick={0.55, 0.57, 0.59, 0.61, 0.63,0.65},
			ylabel near ticks,
			xlabel near ticks,
			tick label style={font=\scriptsize},
			xlabel style={yshift=1ex},
			ylabel style={yshift=-1ex},
		]
		 \addplot [red]
		    table[x=epo, y=p1,] {results/asnq_base.txt};
		\addplot [blue]
		    table[x=epo, y=p1] {results/asnq_tuples.txt};   
	\end{axis}
	\begin{axis}[
	        name=plotCenter1,
			at=(plotLeft1.right of south east), anchor=left of south west,
			scale=\myscale,
			ymin=0.7,
			grid=both,
			xmax=50,
			ytick={.72,.75,.78,.81},
			xtick={0,10,20,30,40},
			xlabel={\# epochs},
			ylabel near ticks,
			xlabel near ticks,
			tick label style={font=\scriptsize},
			xlabel style={yshift=1ex},
		]
%
		
		
		\addplot [green]
		    table[x=epo, y=p1] {results/wikiqa_siamese_tanda.txt};
		\addplot [name path=upper_wst,draw=none] table[x=epo,y expr=\thisrow{p1}+\thisrow{p1_std}] {results/wikiqa_siamese_tanda.txt};
		\addplot [name path=lower_wst,draw=none] table[x=epo,y expr=\thisrow{p1}-\thisrow{p1_std}] {results/wikiqa_siamese_tanda.txt};
		\addplot [fill=green!20] fill between[of=upper_wst and lower_wst];
		    
		\addplot [red]
		    table[x=epo, y=p1,] {results/wikiqa_baseline_tanda.txt};
		\addplot [name path=upper_wbt,draw=none] table[x=epo,y expr=\thisrow{p1}+\thisrow{p1_std}] {results/wikiqa_baseline_tanda.txt};
		\addplot [name path=lower_wbt,draw=none] table[x=epo,y expr=\thisrow{p1}-\thisrow{p1_std}] {results/wikiqa_baseline_tanda.txt};
		\addplot [fill=red!20] fill between[of=upper_wbt and lower_wbt];
		
		\addplot [blue]
		    table[x=epo, y=p1] {results/wikiqa_tuples_tanda.txt};
		\addplot [name path=upper_wtt,draw=none] table[x=epo,y expr=\thisrow{p1}+\thisrow{p1_std}] {results/wikiqa_tuples_tanda.txt};
		\addplot [name path=lower_wtt,draw=none] table[x=epo,y expr=\thisrow{p1}-\thisrow{p1_std}] {results/wikiqa_tuples_tanda.txt};
		\addplot [fill=blue!20] fill between[of=upper_wtt and lower_wtt];
		
	\end{axis}
	\begin{axis}[
	        name=plotRight1,
			at=(plotCenter1.right of south east), anchor=left of south west,
			scale=\myscale,
			ymin=.935,
			grid=both,
			xlabel={\# epochs},
			xmax=20,
			ylabel near ticks,
			xlabel near ticks,
			xtick={0,5,10,15,20},
			tick label style={font=\scriptsize},
			xlabel style={yshift=1ex},
		]

		\addplot [green]
		    table[x=epo, y=p1,] {results/squad_siamese.txt};
		\addplot [name path=upper_ss,draw=none] table[x=epo,y expr=\thisrow{p1}+\thisrow{p1_std}] {results/squad_siamese.txt};
		\addplot [name path=lower_ss,draw=none] table[x=epo,y expr=\thisrow{p1}-\thisrow{p1_std}] {results/squad_siamese.txt};
		\addplot [fill=green!20] fill between[of=upper_ss and lower_ss];
		
		\addplot [red]
		    table[x=epo, y=p1,] {results/squad_baseline.txt};
		\addplot [name path=upper_sb,draw=none] table[x=epo,y expr=\thisrow{p1}+\thisrow{p1_std}] {results/squad_baseline.txt};
		\addplot [name path=lower_sb,draw=none] table[x=epo,y expr=\thisrow{p1}-\thisrow{p1_std}] {results/squad_baseline.txt};
		\addplot [fill=red!20] fill between[of=upper_sb and lower_sb];

		\addplot [blue]
		    table[x=epo, y=p1] {results/squad_tuples.txt};
		\addplot [name path=upper_st,draw=none] table[x=epo,y expr=\thisrow{p1}+\thisrow{p1_std}] {results/squad_tuples.txt};
		\addplot [name path=lower_st,draw=none] table[x=epo,y expr=\thisrow{p1}-\thisrow{p1_std}] {results/squad_tuples.txt};
		\addplot [fill=blue!20] fill between[of=upper_st and lower_st];
		
%
		
		\addplot [black] coordinates {(0,0.9528) (20,0.9528)};
		
	\end{axis}
\end{tikzpicture}
\caption{Local context results (including standard deviation) computed on the dev.~sets.}
\label{fig:res}
\vspace{-1em}
\end{figure}

%% file: charts_global.tex
\begin{figure}[t]
\centering

\resizebox{\linewidth}{!}{
\begin{tikzpicture}
        \begin{customlegend}
        	[legend columns=5,legend style={align=center,column sep=2ex},
        	legend entries={Transformer, \textsc{Loc\_T}, \textsc{Glob\_B}, \textsc{Glob\_E}, \textsc{Dual-CTX}}]
        \addlegendimage{red, line width=1pt}
        \addlegendimage{blue, line width=1pt}
        \addlegendimage{orange, line width=1pt,dashed}
        \addlegendimage{green, line width=1pt,dashed}
        \addlegendimage{cyan, line width=1pt}
        \end{customlegend}
\end{tikzpicture}
}

\begin{tikzpicture}
	\begin{axis}[
	        name=plotLeft,
			scale=\myscale,
			ylabel={MAP},
			ytick={0.6, 0.65, 0.7, 0.75, 0.8},
			grid=both,
			xmax=50,
			xtick={0,10,20,30,40},
			ymin=0.58,
			ylabel near ticks,
			ylabel style={yshift=-1ex},
			tick label style={font=\scriptsize},
			xlabel style={yshift=1ex},
			title={WikiQA},
			title style={yshift=-1.5ex,},
		]
		\addplot [cyan]
		    table[x=epo, y=map,] {results/wikiqa_mixbow.txt};
		\addplot [name path=upper_wbt,draw=none] table[x=epo,y expr=\thisrow{map}+\thisrow{map_std}] {results/wikiqa_mixbow.txt};
		\addplot [name path=lower_wbt,draw=none] table[x=epo,y expr=\thisrow{map}-\thisrow{map_std}] {results/wikiqa_mixbow.txt};
		\addplot [fill=cyan!20] fill between[of=upper_wbt and lower_wbt];
		
		\addplot [green,dashed]
		    table[x=epo, y=map,] {results/wikiqa_globalaver.txt};
		
		\addplot [orange,dashed]
		    table[x=epo, y=map,] {results/wikiqa_globalbow.txt};

		\addplot [red]
		    table[x=epo, y=map,] {results/wikiqa_baseline.txt};
		\addplot [name path=upper_wbt,draw=none] table[x=epo,y expr=\thisrow{map}+\thisrow{map_std}] {results/wikiqa_baseline.txt};
		\addplot [name path=lower_wbt,draw=none] table[x=epo,y expr=\thisrow{map}-\thisrow{map_std}] {results/wikiqa_baseline.txt};
		\addplot [fill=red!20] fill between[of=upper_wbt and lower_wbt];
		
		\addplot [blue]
		    table[x=epo, y=map] {results/wikiqa_tuples.txt};
		    
	\end{axis}
	\begin{axis}[
	        name=plotCenter,
			at=(plotLeft.right of south east), anchor=left of south west,
			scale=\myscale,
			ymin=0.8,
			xtick={0,10,20,30,40},
			xmax=50,
			grid=both,
			ylabel near ticks,
			tick label style={font=\scriptsize},
			xlabel style={yshift=1ex},
			title={WikiQA (TANDA)},
			title style={yshift=-1.5ex,},
		]
%

		\addplot [cyan]
		    table[x=epo, y=map] {results/wikiqa_mixbow_tanda.txt};
		\addplot [name path=upper_wtt,draw=none] table[x=epo,y expr=\thisrow{map}+\thisrow{map_std}] {results/wikiqa_mixbow_tanda.txt};
		\addplot [name path=lower_wtt,draw=none] table[x=epo,y expr=\thisrow{map}-\thisrow{map_std}] {results/wikiqa_mixbow_tanda.txt};
		\addplot [fill=cyan!20] fill between[of=upper_wtt and lower_wtt];

		
		\addplot [red]
		    table[x=epo, y=map,] {results/wikiqa_baseline_tanda.txt};
		\addplot [name path=upper_wbt,draw=none] table[x=epo,y expr=\thisrow{map}+\thisrow{map_std}] {results/wikiqa_baseline_tanda.txt};
		\addplot [name path=lower_wbt,draw=none] table[x=epo,y expr=\thisrow{map}-\thisrow{map_std}] {results/wikiqa_baseline_tanda.txt};
		\addplot [fill=red!20] fill between[of=upper_wbt and lower_wbt];
		
		\addplot [blue]
		    table[x=epo, y=map] {results/wikiqa_tuples_tanda.txt};
		\addplot [name path=upper_wtt,draw=none] table[x=epo,y expr=\thisrow{map}+\thisrow{map_std}] {results/wikiqa_tuples_tanda.txt};
		\addplot [name path=lower_wtt,draw=none] table[x=epo,y expr=\thisrow{map}-\thisrow{map_std}] {results/wikiqa_tuples_tanda.txt};
		\addplot [fill=blue!20] fill between[of=upper_wtt and lower_wtt];
	\end{axis}
	\begin{axis}[
	        name=plotRight,
			at=(plotCenter.right of south east), anchor=left of south west,
			scale=\myscale,
			ymin=.945,
			grid=both,
			xtick={0,5,10,15,20},
			xmax=20,
			ylabel near ticks,
			tick label style={font=\scriptsize},
			xlabel style={yshift=1ex},
			title={SQuAD},
			title style={yshift=-1.5ex,},
		]
		
		\addplot [cyan]
		    table[x=epo, y=map,] {results/squad_mixbow.txt};
		
		\addplot [green,dashed]
		    table[x=epo, y=map,] {results/squad_globalaver.txt};
		
		\addplot [orange,dashed]
		    table[x=epo, y=map,] {results/squad_globalbow.txt};

		\addplot [red]
		    table[x=epo, y=map,] {results/squad_baseline.txt};
		\addplot [name path=upper_wbt,draw=none] table[x=epo,y expr=\thisrow{map}+\thisrow{map_std}] {results/squad_baseline.txt};
		\addplot [name path=lower_wbt,draw=none] table[x=epo,y expr=\thisrow{map}-\thisrow{map_std}] {results/squad_baseline.txt};
		\addplot [fill=red!20] fill between[of=upper_wbt and lower_wbt];
		
		\addplot [blue]
		    table[x=epo, y=map] {results/squad_tuples.txt};
	\end{axis}
\end{tikzpicture}


\begin{tikzpicture}
	\begin{axis}[
	        	name=plotLeft1,
	        	scale=\myscale,
			xlabel={\# epochs},
			ylabel={P@1},
			grid=both,
			xtick={0,10,20,30,40},
			xmax=50,
			ytick={0.55, 0.6, 0.65, 0.7},
			ymin=0.52,
			ylabel near ticks,
			xlabel near ticks,
			ylabel style={yshift=-1ex},
			tick label style={font=\scriptsize},
			xlabel style={yshift=1ex},
		]
		\addplot [cyan]
		    table[x=epo, y=p1,] {results/wikiqa_mixbow.txt};
		\addplot [name path=upper_wbt,draw=none] table[x=epo,y expr=\thisrow{p1}+\thisrow{p1_std}] {results/wikiqa_mixbow.txt};
		\addplot [name path=lower_wbt,draw=none] table[x=epo,y expr=\thisrow{p1}-\thisrow{p1_std}] {results/wikiqa_mixbow.txt};
		\addplot [fill=cyan!20] fill between[of=upper_wbt and lower_wbt];
		
		\addplot [green,dashed]
		    table[x=epo, y=p1,] {results/wikiqa_globalaver.txt};
		
		\addplot [orange,dashed]
		    table[x=epo, y=p1,] {results/wikiqa_globalbow.txt};

		\addplot [red]
		    table[x=epo, y=p1,] {results/wikiqa_baseline.txt};
		\addplot [name path=upper_wbt,draw=none] table[x=epo,y expr=\thisrow{p1}+\thisrow{p1_std}] {results/wikiqa_baseline.txt};
		\addplot [name path=lower_wbt,draw=none] table[x=epo,y expr=\thisrow{p1}-\thisrow{p1_std}] {results/wikiqa_baseline.txt};
		\addplot [fill=red!20] fill between[of=upper_wbt and lower_wbt];
		
		\addplot [blue]
		    table[x=epo, y=p1] {results/wikiqa_tuples.txt};
	\end{axis}
	\begin{axis}[
	        name=plotCenter1,
			at=(plotLeft1.right of south east), anchor=left of south west,
			scale=\myscale,
			ymin=0.7,
			grid=both,
			xmax=50,
			ytick={.72,.75,.78,.81},
			xtick={0,10,20,30,40},
			xlabel={\# epochs},
			ylabel near ticks,
			xlabel near ticks,
			tick label style={font=\scriptsize},
			xlabel style={yshift=1ex},
		]
%
		
		\addplot [cyan]
		    table[x=epo, y=p1] {results/wikiqa_mixbow_tanda.txt};
		\addplot [name path=upper_wtt,draw=none] table[x=epo,y expr=\thisrow{p1}+\thisrow{p1_std}] {results/wikiqa_mixbow_tanda.txt};
		\addplot [name path=lower_wtt,draw=none] table[x=epo,y expr=\thisrow{p1}-\thisrow{p1_std}] {results/wikiqa_mixbow_tanda.txt};
		\addplot [fill=cyan!20] fill between[of=upper_wtt and lower_wtt];
		
		    
		\addplot [red]
		    table[x=epo, y=p1,] {results/wikiqa_baseline_tanda.txt};
		\addplot [name path=upper_wbt,draw=none] table[x=epo,y expr=\thisrow{p1}+\thisrow{p1_std}] {results/wikiqa_baseline_tanda.txt};
		\addplot [name path=lower_wbt,draw=none] table[x=epo,y expr=\thisrow{p1}-\thisrow{p1_std}] {results/wikiqa_baseline_tanda.txt};
		\addplot [fill=red!20] fill between[of=upper_wbt and lower_wbt];
		
		\addplot [blue]
		    table[x=epo, y=p1] {results/wikiqa_tuples_tanda.txt};
		\addplot [name path=upper_wtt,draw=none] table[x=epo,y expr=\thisrow{p1}+\thisrow{p1_std}] {results/wikiqa_tuples_tanda.txt};
		\addplot [name path=lower_wtt,draw=none] table[x=epo,y expr=\thisrow{p1}-\thisrow{p1_std}] {results/wikiqa_tuples_tanda.txt};
		\addplot [fill=blue!20] fill between[of=upper_wtt and lower_wtt];
		
	\end{axis}
	\begin{axis}[
	        name=plotRight1,
			at=(plotCenter1.right of south east), anchor=left of south west,
			scale=\myscale,
			ymin=.935,
			grid=both,
			xlabel={\# epochs},
			xmax=20,
			ylabel near ticks,
			xlabel near ticks,
			xtick={0,5,10,15,20},
			tick label style={font=\scriptsize},
			xlabel style={yshift=1ex},
		]

		\addplot [cyan]
		    table[x=epo, y=p1,] {results/squad_mixbow.txt};
		
		\addplot [green,dashed]
		    table[x=epo, y=p1,] {results/squad_globalaver.txt};
		
		\addplot [orange,dashed]
		    table[x=epo, y=p1,] {results/squad_globalbow.txt};

		\addplot [red]
		    table[x=epo, y=p1,] {results/squad_baseline.txt};
		\addplot [name path=upper_wbt,draw=none] table[x=epo,y expr=\thisrow{p1}+\thisrow{p1_std}] {results/squad_baseline.txt};
		\addplot [name path=lower_wbt,draw=none] table[x=epo,y expr=\thisrow{p1}-\thisrow{p1_std}] {results/squad_baseline.txt};
		\addplot [fill=red!20] fill between[of=upper_wbt and lower_wbt];
		
		\addplot [blue]
		    table[x=epo, y=p1] {results/squad_tuples.txt};
		
		\addplot [black] coordinates {(0,0.9528) (20,0.9528)};
		
	\end{axis}
\end{tikzpicture}
\caption{Global context - empirical results computed on the development sets. The standard deviation is not always exposed to improve the readability.}
\vspace{-.7em}
\label{fig:res_global}
\end{figure}
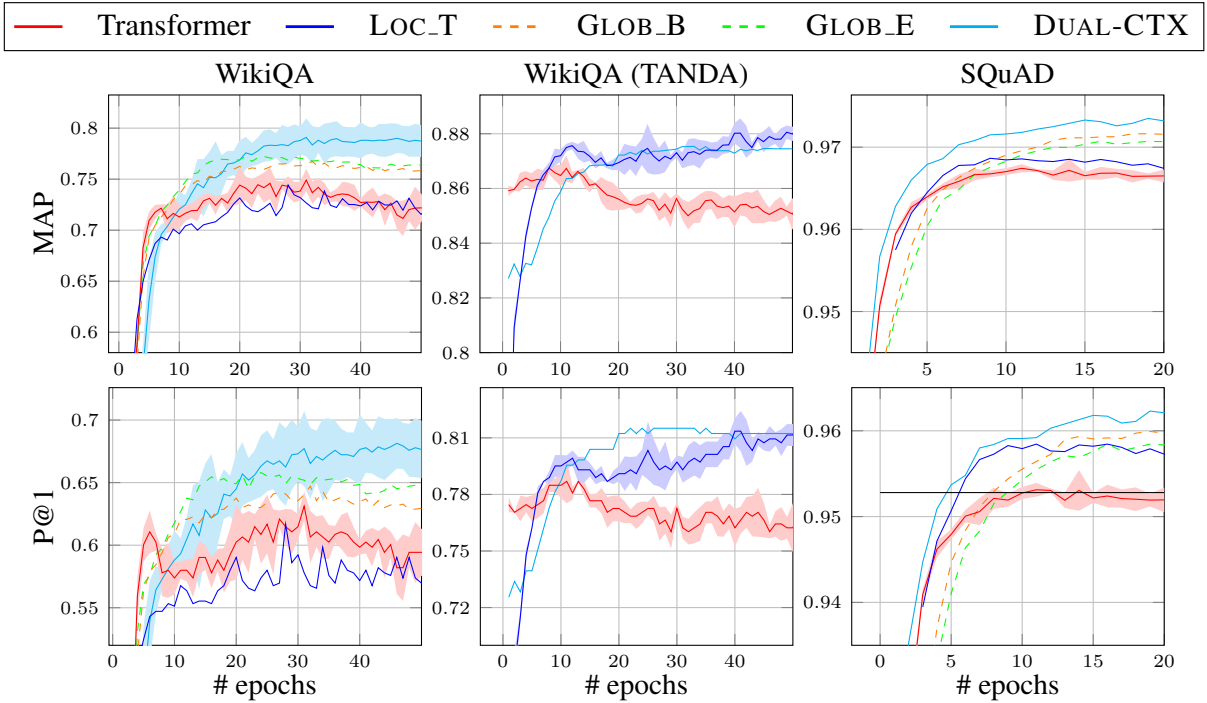